\documentclass[runningheads]{llncs}
\usepackage{graphicx}
\usepackage{times}
\usepackage{latexsym}
\usepackage[T1]{fontenc}
\usepackage[utf8]{inputenc}
\usepackage{microtype}
\usepackage{inconsolata}
\usepackage{diagbox}
\usepackage{graphicx}
\usepackage[table]{xcolor}
\usepackage{makecell}
\usepackage{multirow}
\usepackage{booktabs}
\usepackage{listings}
\usepackage{svg}
\usepackage{pdfpages}
\usepackage{times}
\usepackage{latexsym}
\usepackage[T1]{fontenc}
\usepackage[utf8]{inputenc}
\usepackage{microtype}
\usepackage{inconsolata}
\usepackage{diagbox}
\usepackage{graphicx}
\usepackage[table]{xcolor}
\usepackage{makecell}
\usepackage{multirow}
\usepackage{booktabs}
\usepackage{listings}
\usepackage{svg}
\usepackage{cite}
\usepackage{pdfpages}
\begin{document}
\title{Empowering Refugee Claimants and their Lawyers: Using Machine Learning to Examine Decision-Making in Refugee Law}
\titlerunning{Using Machine Learning to Examine Decision-Making in Refugee Law}

%
\author{Claire Barale\inst{1}\orcidID{0000-0002-9798-0716}}
\authorrunning{Claire Barale}
\institute{The University of Edinburgh, School of Informatics \\
\email{claire.barale@ed.ac.uk}\\}
\maketitle              
\section{Introduction}
Legal NLP is an active and promising field of research. NLP application in the legal profession also presents significant challenges for researchers and legal professionals. A wide range of functionalities have been explored in legal NLP \cite{zhong-etal-2018-legal}, such as legal summarization \cite{elaraby-litman-2022-arglegalsumm, aumiller-etal-2022-eurlex}, legal search and legal information extraction and retrieval \cite{bommarito2021lexnlp, bruninghaus2001improving}, automatic text generation, text classification \cite{chalkidis-etal-2022-fairlex, chalkidis-etal-2019-large} or legal prediction \cite{dunn_early_2017, katz_general_2017, kaur_convolutional_nodate, branting_semi-supervised_2019, pagallobranting2018inducing, chen_asylum_2017}. While there has been extensive research in the broader field of NLP, there are fewer studies in the field of refugee law, and at the time of writing this paper, no significant contribution has been based on raw decision reports.

Our project aims at helping and supporting stakeholders in refugee status adjudications, such as lawyers, judges, governing bodies, and claimants, in order to make better decisions through data-driven intelligence and increase the understanding and transparency of the refugee application process for all involved parties. This PhD project has two primary objectives: (1) to retrieve past cases, and (2) to analyze legal decision-making processes on a dataset of Canadian cases. In this paper, we present the current state of our work, which includes a completed experiment on part (1) and ongoing efforts related to part (2). We believe that NLP-based solutions are well-suited to address these challenges, and we investigate the feasibility of automating all steps involved. In addition, we introduce a novel benchmark for future NLP research in refugee law. Our methodology aims to be inclusive to all end-users and stakeholders \cite{barale-2022-human}, with expected benefits including reduced time-to-decision, fairer and more transparent outcomes, and improved decision quality.

Our contributions are as follows:
\begin{enumerate}
    \item {Creating a novel dataset in refugee law with gold-standard and silver-standard annotations to be used as a benchmark in legal NLP}
    \item {Building an end-to-end pipeline for a global extraction tool via NER models, which extracts targeted information from raw text and gives structure to the cases (achieved)}
    \item {Developing a tool for legal search based on the structured dataset and SQLite database, which enables legal professionals to retrieve past cases quickly and easily (achieved)}
    \item {Performing a prediction experiment to analyze judgment consistency based on the raw text and the unstructured and structured dataset, using state-of-the-art explainability methods (in progress: preliminary results)}
    \item {Conducting a survey on HCI and NLP to examine the use of the tool by legal professionals (preliminary survey achieved)}
\end{enumerate}

\section{Background and motivation}

At the heart of the ongoing refugee crisis lies the legal and administrative process known as \textit{Refugee Status Determination} (\textit{RSD}), which can be explained into three distinct sub-procedures: (i) the formal submission of a refugee protection claim by an applicant, often assisted by legal counsel, (ii) the decision-making process conducted by a panel of judges, and (iii) the ultimate decision outcome accompanied by a written rationale for either granting or denying refugee protection. These decisions bear significant implications and affect approximately 4.6 million asylum seekers worldwide as of mid-2022. In the case of Canada in 2021, there were 48,014 new claims and 10,055 appeals filed \footnote{\url{https://irb.gc.ca/en/statistics/Pages/index.aspx}}. The processing times for refugee claims can vary widely, spanning from a few months to several years. One contributing factor to these extended processing times is the extensive effort required for conducting research on similar cases, a critical aspect of the counsel's preparatory work for new claim files.

The goal of the first part of this project is to support legal practitioners, both lawyers preparing the application file and judges reasoning the decision outcome, by automating the time-consuming search for similar legal cases known here as \textit{refugee case analysis}. A common approach used by legal practitioners is to manually search and filter past \textit{RSD} cases on online databases such as CanLII or Refworld by elementary \textit{document text} search. There are several expected benefits for legal practitioners: (i) to facilitate, speed up and specify legal search, (ii) for lawyers, to reduce the time spent on a claim, and to provide relevant references, potentially resulting in a file that has more chances of being accepted, and (iii) for judges, to remain consistent across time. Building on this study, we work on analyzing the decision-making process. Based on the structured dataset created, we highlight the correlation between a range of features in the form of tokens and the final decision of the case. 

\section{Research approach}
The goal is to determine whether state-of-the-art methods can be used to improve the transparency and processing of refugee cases. Transparent, high-quality decisions are understood to be well-informed, traceable, and reproducible. The general interrogation driving the PhD work is: \textbf{Can state-of-the-art natural language processing methods be used to improve transparency and quality of the asylum decision-making process in collaboration with humans?}

Key research questions include:

\noindent\textbf{Training data requirements}  How many labeled samples are needed, both gold and silver standard? 

\noindent\textbf{Information extraction} What methods are best suited to identify and extract the targeted information from legal cases? 

\noindent\textbf{Replicability}  How generalizable could this work be on other legal data sets (other legal fields or other jurisdictions)?

\noindent\textbf{Pre-training}  How important is the domain match: are domain-specific pre-training able to perform better than general-purpose embeddings, despite their smaller size?
    
\noindent\textbf{Architectures}  How important is the architecture as applied to the information extraction tasks, in terms of F1-score, precision, and recall?

\noindent\textbf{Judgment prediction} Can a binary classification task using neural networks and large language models uncover biases and lack of consistency in the outcomes of the decisions? 

\noindent\textbf{Explainability} What methods can we use to highlight those inconsistencies? Are state-of-the-art posthoc explainable methods appropriate to our use case? Can we measure causality instead of correlation?

\section{Data set and data characteristics: a novel benchmark dataset for refugee law}
\subsection{Information retrieval}
We retrieve 59,112 historic decision documents (dated 1996 to 2022) from online services of the Canadian Legal Information Institute (CanLII) based on context-based indexing and metadata to curate a collection of federal refugee \textit{RSD} cases. Our automated retrieval process is exhaustive and comprises all available cases. It is superior to human-based manual retrieval in terms of error proneness and processing time. We obtain two sets: (1) a set of \textit{case covers} that consists of semi-structured data and displays meta-information and (2) a set of \textit{main text} that contains the body of each case, in full text.

\subsection{Collecting gold-standard annotated data}
While this is a time-consuming process, it is very important to have high-quality annotations. Thus, we chose to have gold-standard annotation.  In order to speed up the process we used state-of-the-art semi-automatic annotation tools. We use \texttt{Prodigy} annotation tool\footnote{\texttt{Prodigy}: \url{https://prodi.gy/docs}} using an academic research license in order to speed up and improve the manual labeling work in terms of consistency and accuracy of annotations. We annotate 346 case covers, and 2,436 sentences for the main text, randomly chosen.

\begin{figure*}
   \includegraphics[width=\textwidth]{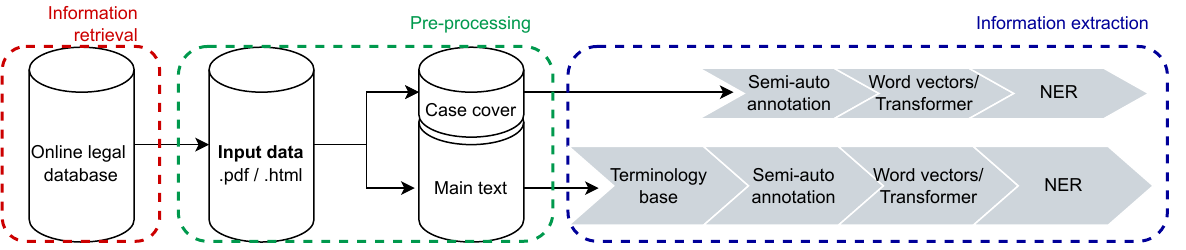} 
   \centering
   \caption{End-to-end automated pipeline}
   \label{figure: pipeline}
\end{figure*}

To collect annotated samples for conventional NER labels (\texttt{DATE, ORG, GPE, PERSON, NORP, LAW}), we leverage suggestions from general-purpose pretrained embeddings \footnote{\url{https://spacy.io/models/en}}. For the remaining labels (\texttt{CLAIMANT\_INFO, CLAIMANT\_EVENT, PROCEDURE, DOC\_EVIDENCE, EXPLANATION, DETERMINATION, CREDIBILITY}) and in order to enhance annotation consistency, we establish a terminology repository (Figure \ref{figure: pipeline}), relying on \texttt{word2vec} \cite{mikolov2013efficient}. During the human annotation task, patterns are cross-referenced with displayed sentences, with human annotators primarily responsible for rectifying any discrepancies. This approach yields a meticulously annotated set of sentences, significantly expediting the labeling process. Table \ref{table: target items} provides a breakdown of the labels present in our annotated dataset.

\subsection{Generating silver-standard annotated data}
\begin{figure}
\centering
\begin{minipage}{.5\textwidth}
  \centering
  \includegraphics[width=.8\linewidth]{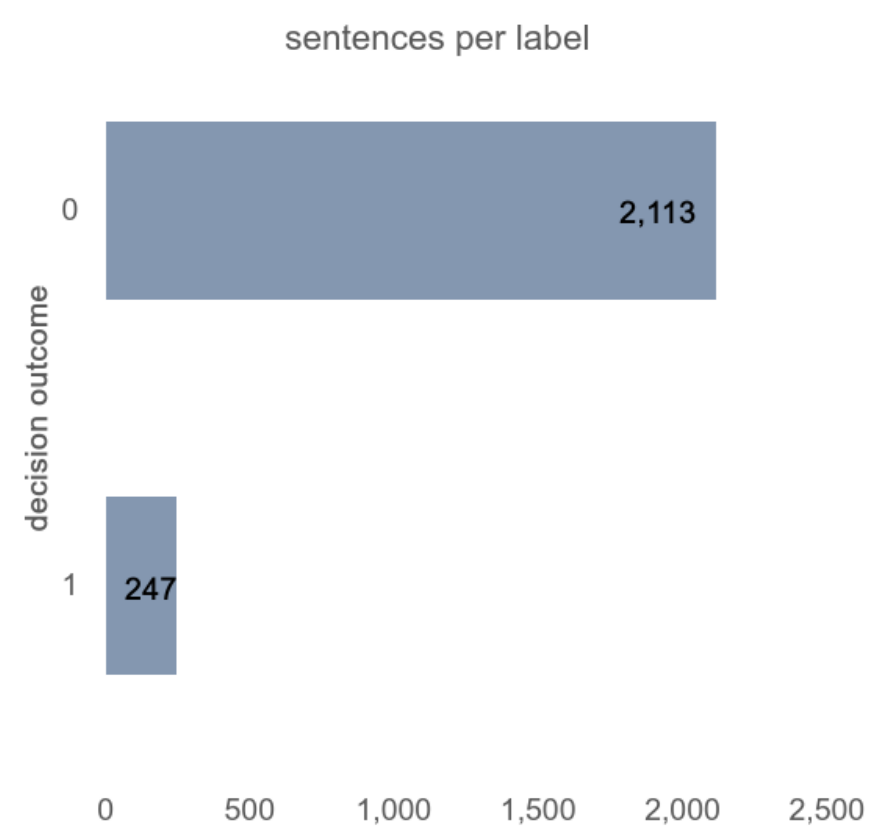} 
  \caption{Distribution of gold-standard \textcolor{white}{xxxxxxxxxx} annotations for case outcome}  
  \label{fig:gold}
\end{minipage}%
\begin{minipage}{.5\textwidth}
  \centering
  \includegraphics[width=.8\linewidth]{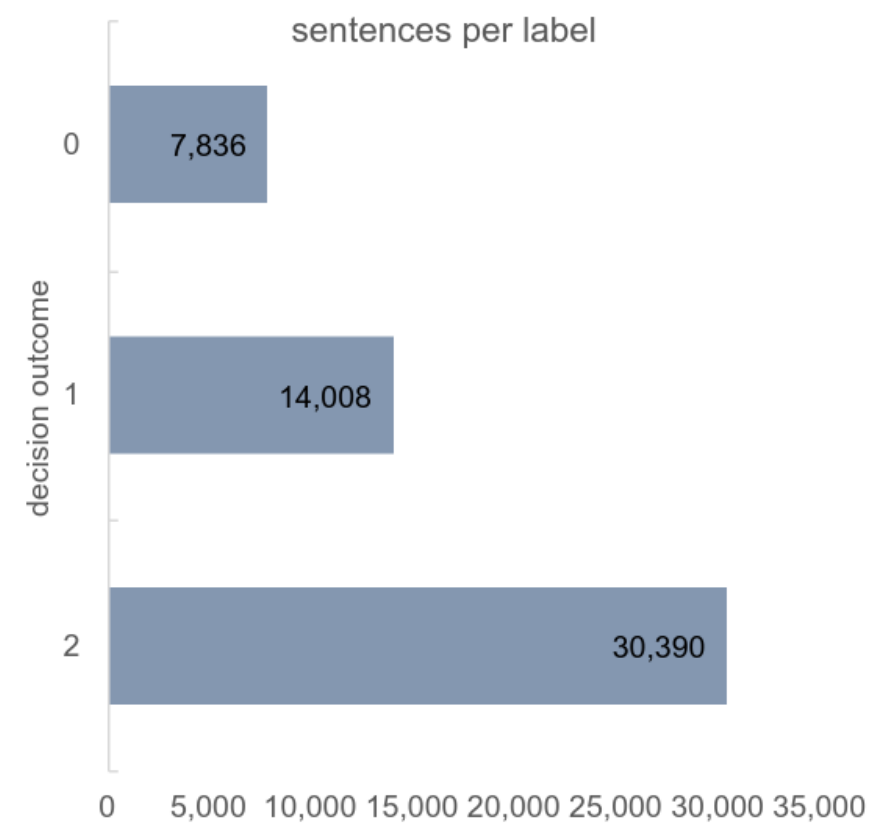} 
  \caption{Distribution of silver-standard annotations for case outcome}
  \label{fig:silver}
\end{minipage}
\end{figure}

After annotating our dataset with gold-standard labels, we used a transformer-based text classification model (BERT) to generate silver-standard labels that indicate the outcome of each case. To accomplish this, we first used our NER model (described in Section \ref{NER}) to extract the sentences pertaining to the decision outcomes, which were identified by the NER with the label "Determination". We then trained a classifier on 2,360 labeled sentences with positive or negative outcomes (Fig. \ref{fig:gold} showing 10.47\% of claims granted with the label 1 and 89.53\% of rejected claims with the label 0), using all of the extracted sentences. Since there may be multiple extracted sentences per case, we employed a majority vote mechanism to determine the outcome of each case. Sentences that could not be confidently classified as positive or negative (with a computed weight between 0.4 and 0.6) were categorized as 'Uncertain'. Our classifier achieved 90\% accuracy, classifying 52,234 sentences into three categories: granted (1), denied (0), and uncertain (2) (Fig. \ref{fig:silver}, showing 58.18\% of claims denied, 26.82\% of claims granted and 15\% uncertain). In total, we obtained decision outcome labels for approximately 25,000 cases. Similarly, running our previously trained NER models on the remaining data allows us to collect silver-standard annotations for the whole dataset (as explained below in section \ref{NER}).

\begin{table*}[h!]
\centering
\resizebox{\textwidth}{!}{
\begin{tabular}{|c|c|c|c|c|}
\hline
 & \textbf{Label} & \textbf{A} & \textbf{Description} & \textbf{Example} \\

\hline

 \multirow{5}{*}{\rotatebox[origin=c]{90}{\textbf{Case cover}}} & \multicolumn{4}{c|}{\textbf{\textit{General}}} \\
 \cline{2-5}
 & \texttt{DATE} & 1,219 & absolute or relative dates or periods & date of the hearing and date of the decision \\
 & \texttt{GPE} & 871 & cities, countries, regions & place of the hearing\\
 & \texttt{ORG} & 278 & tribunals & "immigration appeal division", "refugee protection division" \\
 & \texttt{PERSON} & 119 & names & name of the panel and counsels\\
 \hline 
 \multirow{20}{*}{\rotatebox[origin=c]{90}{\textbf{Main text}}} &\multicolumn{4}{c|}{\textbf{\textit{Information on claimant and allegations}}} \\
 \cline{2-5}
 & \texttt{CLAIMANT\_EVENT} & 1,575 & verbs or nouns describing an event of the story of the claimant & "rape", "threat", "attacks", "fled"\\
 & \texttt{CLAIMANT\_INFO} & 235 & age, gender, citizenship, occupation & "28 year old", "citizen of Iran", "female" \\
 & \texttt{GPE} & 732 & cities, countries, regions & countries of past residency or places of hearings: "toronto, ontario" \\
 & \texttt{NORP} & 129 & nationalities, religious, political or ethnic groups or communities & "hutu", "nigerian", "christian" \\
 \cline{2-5}
 & \multicolumn{4}{c|}{\textbf{\textit{Legal procedure}}} \\
 \cline{2-5}
 & \texttt{ORG} & 549 & tribunals, NGOs, companies & "human rights watch", "refugee protection division" \\
 & \texttt{PROCEDURE} & 594 & steps in the claim and legal procedure events & "removal order", "sponsorship for application" \\
 \cline{2-5}
 & \multicolumn{4}{c|}{\textbf{\textit{Analysis and reasons for decision outcome}}} \\
 \cline{2-5}
 & \texttt{CREDIBILITY} & 684 & mentions of credibility in the determination & "lack of evidence", "inconsistencies" \\
 & \texttt{DETERMINATION} & 76 & outcome of the decision (accept/reject) & "appeal is dismissed", "panel determines that the claimant is not a convention refugee" \\
 & \texttt{DOC\_EVIDENCE} & 768 & evidences, proofs, supporting documents & "passport", "medical record", "marriage certificate" \\
 & \texttt{EXPLANATION} & 404 & reasons given by the panel for the determination & "fear of persecution", "no protection by the state" \\
 \cline{2-5}
 & \multicolumn{4}{c|}{\textbf{\textit{Timeline}}} \\
 \cline{2-5}
 & \texttt{DATE} & 628 & absolute or relative dates or periods  & "for two ears", "june, 4th 1996" \\
 \cline{2-5}
 & \multicolumn{4}{c|}{\textbf{\textit{Names}}}\\
 \cline{2-5}
 & \texttt{PERSON} & 154 & names & claimants' names, their family, name of judges\\
 \cline{2-5}
 & \multicolumn{4}{c|}{\textbf{\textit{Citations}}} \\
 \cline{2-5} 
 & \texttt{LAW} & 476 & legislation and international conventions & state law and international conventions, "section 1(a) of the convention" \\
 & \texttt{LAW\_CASE} & 109 & case law and past decided cases, by the same tribunal or another & "xxx v. minister of canada, 1994" \\
 & \texttt{LAW\_REPORT} & 18 & country reports written by NGOs or the United Nations & " 	amnesty international, surviving death: police and military torture of women in mexico, 2016" \\

\hline

\end{tabular}}
\caption{Targeted categories (\textbf{Label}) for extraction with number of annotations (\textbf{A}) per label, sorted alphabetically.}
\label{table: target items}
\end{table*}

\section{Named-Entity Recognition task: a structured dataset for past cases retrieval} \label{NER}
The task of past case retrieval and historical case analysis can be challenging for legal professionals due to the sheer volume of documents to be searched, which can be time-consuming and costly. Existing tools for past case retrieval provide only document-level solutions, which can lack precision and transparency. Our proposed solution addresses these issues by offering a higher level of precision and enabling users to select relevant criteria, resulting in greater transparency and efficiency.

In our research, we ues neural network techniques for information extraction from legal documents. Conventional methods, such as pattern matching and regular expressions, proved overly restrictive for our diverse dataset. Furthermore, we opted against unsupervised approaches to ensure transparency and allow legal professionals to define their own similarity criteria. Instead, we leveraged sequence-labeling classification and Named Entity Recognition (NER) methods to capture keywords and concise phrases. We curated a training dataset comprising annotated examples for each specific category of interest and fine-tuned state-of-the-art neural network models using our collection of legal cases. Ultimately, we extracted targeted items from the cases related to the specified categories and organized them into a structured database.

The chosen categories of interest are described in Table \ref{table: target items}. These labels were crafted and finalized through collaboration with three refugee lawyers. Drawing from interviews with these legal experts, we compiled a comprehensive inventory of keywords, grounds, and legal components that significantly influence decision-making. Furthermore, we conducted an in-depth analysis of a sample comprising 50 cases, thoughtfully recommended by the interviewees to span a representative spectrum of claims and tribunals over the years.

We trained the NER models using 80\% of the labeled data as a training set (respectively 276 case covers and 1,951 sentences for the main text), 10\% of the labeled data as a development set (35 case covers and 244 sentences), and 10\% of the labeled data as test set for evaluation (35 case covers and 244 sentences).

\subsection{Architectures}

We conducted experiments with five distinct model architectures for the \textit{case cover} and seven different architectures for the \textit{main text}. Specifically, we explored five variants based on Convolutional Neural Networks (CNNs) utilizing diverse word embeddings and two transformers architectures. We use a CNN without added vectors as a baseline. Only the transformer architectures required GPU training. For our experiments, we used the \texttt{spaCy} pipelines\footnote{\texttt{SpaCy}: \url{https://spacy.io/api/entityrecognizer}} (comprising a tokenizer, CNN, and transformers) and leveraged the \texttt{HuggingFace} models \footnote{\texttt{RoBERTa}: \url{https://huggingface.co/roberta-base}, \newline \texttt{LegalBERT}: \url{https://huggingface.co/nlpaueb/legal-bert-base-uncased}}. All CNNs were optimized using the \texttt{Adam} optimizer function. Given the suitability of the masked language modeling objective for sentence-labeling tasks, we conducted experiments with \texttt{roBERTa} \cite{liu2019roberta} and \texttt{LegalBERT} \cite{chalkidis-etal-2020-legal} to compare a general content model with a legal content model.

To serve as the first layer in our NER network, we incorporate pretrained character-level embeddings. This allows us to isolate the impact of pretraining from the effect of the architecture and improve the F1 score for target items. We fine-tuned \texttt{GloVe} vectors (\cite{pennington2014glove}) in 50 dimensions using the \texttt{Mittens}\footnote{\texttt{Mittens}: \url{https://github.com/roamanalytics/mittens}} Python package \cite{dingwall-potts-2018-mittens}, resulting in the creation of 970 static vectors. On top of the generated static vectors, we add dynamic contextualized vectors using pre-training embeddings based on \texttt{BERT} \cite{devlin2019bert}, updating weights on our corpus of cases. Given that the text within the \textit{case cover} is presented in a semi-structured format, we consider that it is unnecessary to perform pre-training.

\subsection{Results} 
\begin{figure*}
   \includegraphics[width=\textwidth]{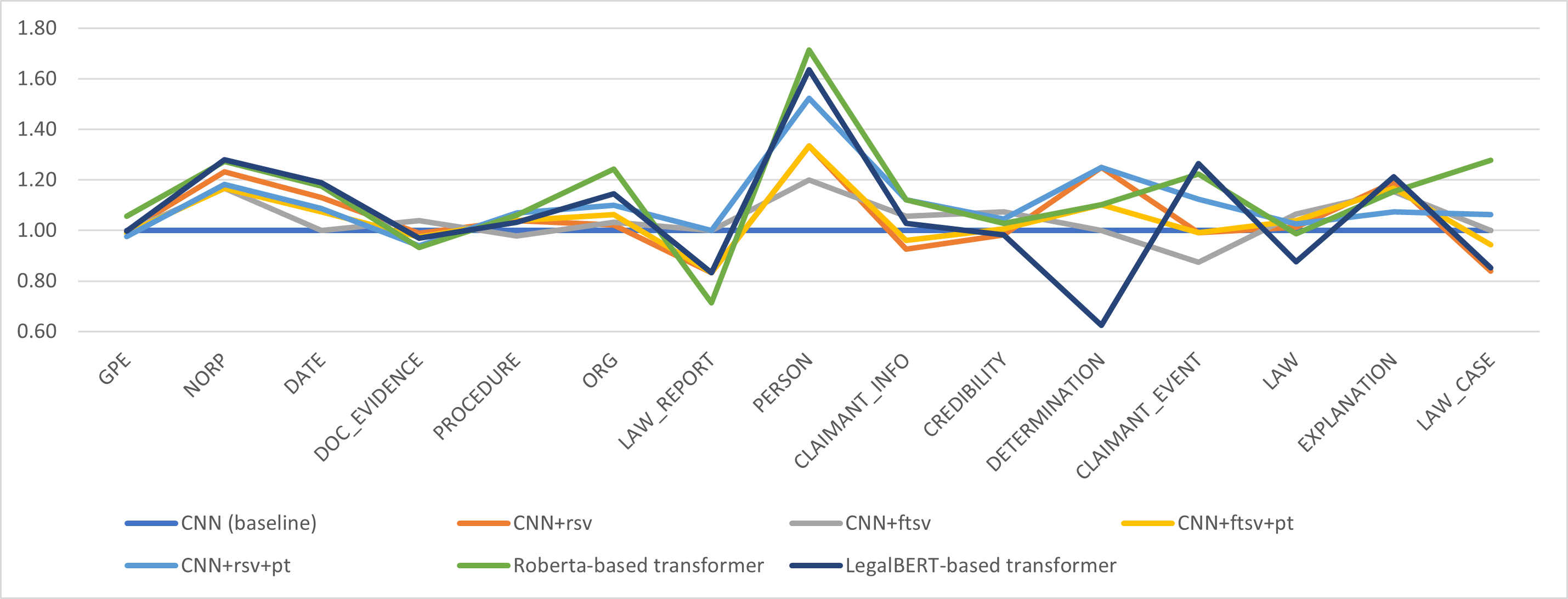} 
   \centering
   \caption{Results on the \textit{main text}: F1-score (in \%, on the y-axis) per targeted category (x-axis) on seven network architectures baseline CNN model (\texttt{baseline}), CNN model with random static vectors on \texttt{en\_core\_web\_lg} (\texttt{CNN+rsv}), CNN with fine-tuned static vectors (\texttt{CNN+fts}), CNN with random static vectors and pre-training (\texttt{CNN+rsv+pt}), CNN with fine-tuned static vectors and pre-training (\texttt{CNN+fts+pt}), \texttt{RoBERTa}-based transformer, \texttt{LegalBERT}-based transformer}
   \label{figure: graph baseline}
\end{figure*}

Results can be visualized in figure \ref{figure: graph baseline} by comparison to our baseline, and are presented in detail in Annex.

For the \textit{case cover}, we achieved very satisfactory results across all labels, attaining an F1-score exceeding 90\% for three of them and reaching 84.78\% for name extraction. For all labels except names, CNN architectures outperformed other approaches, with dates achieving the highest score when initialized with random embeddings. We attribute this phenomenon to the specific page layout of this section. The primary advantage of employing a transformer-based model was observed in achieving a higher recall compared to the CNN-based architectures.

For the \textit{main text}, our results show variability across labels. We achieved scores exceeding 80\% for labels such as \texttt{DATE, GPE, PERSON, ORG}, with the highest performance observed when utilizing \texttt{roBERTa}. However, for labels \texttt{EXPLANATION, LAW, LAW\_CASE}, our scores fell below 60\%. Overall, when employing transformer models, we noted an improved balance between precision and recall. As for labels \texttt{DETERMINATION, LAW\_REPORT, NORP}, the results are deemed unreliable due to the limited size of both the training and testing datasets, thus we choose to not further comment on them.

Across a majority of categories, results show the significance of the pre-training content, particularly for labels such as \texttt{CLAIMANT\_EVENT, CREDIBILITY, DATE, \newline DOC\_EVIDENCE, EXPLANATION, LAW, PROCEDURE}. Our findings indicate that the impact of domain-specific training data outweighs the differences in network architecture. Specifically, it seems that for certain categories (\texttt{CREDIBILITY, DOC\_EVIDENCE, LAW, PROCEDURE}), pre-training on our own dataset yields more effective results than training on a general legal dataset like \texttt{LegalBERT}. This can be attributed to the fact that \texttt{LegalBERT} lacks exposure to Canadian text and refugee cases, as it was trained on data from the US, Europe, and the UK. Conversely, for other categories, \texttt{roBERTa} outperforms both \texttt{LegalBERT} and CNNs, suggesting that the model size of the pre-trained model carries more weight than domain-match considerations. Notably, with \texttt{LegalBERT} being 12GB and \texttt{roBERTa} 160GB, the latter excels and outperforms \texttt{LegalBERT} across traditional NER labels (\texttt{GPE, ORG, PERSON}), as well as \texttt{CLAIMANT\_INFO} and \texttt{LAW\_CASE}.

\subsection{Related work}

Earlier methods for statistical information extraction in the field of law employed linear models such as maximum entropy models \cite{bender2003maximum, clark2003combining} and hidden Markov models \cite{mayfield2003named}. Recent advancements in the field have been achieved by methods capable of capturing contextual information. There exists an active research community exploring the application of conditional random fields \cite{benikova2015c, faruqui2010training, finkel2005incorporating} and Bidirectional Long Short-Term Memory networks (BiLSTMs) \cite{chiu2016named, huang2015bidirectional, lample-etal-2016-neural, ma-hovy-2016-end, leitner2019fine} for various legal applications. The introduction of deep learning architectures, including Recurrent Neural Networks (RNNs), Convolutional Neural Networks (CNNs), and attention mechanisms, has further improved the scope and performance in the field, as demonstrated in \cite{chalkidis-etal-2019-extreme}. However, it's worth noting that transformers do not consistently outperform other models on our dataset. In our work, we focus on statistical approaches to NER.

Matching similar cases is a widely recognized application of NLP techniques in the legal domain, particularly in common law systems \cite{trappey_identify_2020} and domains such as international law. The \textit{Competition on Legal Information Extraction/Entailment} includes a case retrieval task, underscoring the significance of this area in both research and practical applications. While previous research has explored case matching at the paragraph level \cite{tang-clematide-2021-searching, hu2022bert_lf}, our approach prioritizes transparency and empowers legal practitioners to make choices about "what defines similarity" by allowing them to select filters and criteria tailored to their specific needs.

\section{Legal decision-making analysis: unwanted variability in refugee status adjudications?}
Our objective is to determine which tokens or strings of text are the most significant in predicting the outcome of cases.

\subsection{Predictive analysis}
To conduct the analysis, the primary task is to predict the outcome of a case by utilizing the structured dataset. An additional pre-processing step was performed on the extracted categories, which have now been expanded to include the date of the decision, date of the hearing, tribunal, name of the judge, sequence of events (allegations), gender, age, citizenship of the claimant, dependent applicants, single or multiple applicants, mentions of credibility, hard documentary evidence provided, legal procedure events, explanations given by the panel, virtual or in-person hearing, public or private hearing, determination of the case, and legal citations (convention, national law, cases, reports). These features are presented in a string format which are concatenated to create one string per case with a separation token, used as an input for a prediction model. The objective of this second experiment is to establish correlations between the features and the decision outcome.

\subsubsection{Noise and biases in legal decision making}
Noise refers to a random irregularity within a sample. Noise in decision-making is defined as an ``unwanted variability" \cite{sunstein2021governing} that has consequences as it can produce errors of judgment and inconsistency in the outcome of the decisions. Noise is difficult to detect and correct as there are no recognizable patterns in errors (unlike errors that derive from bias). Three kinds of noise can be distinguished: ``Occasion noise" (decisions can be subject to external factors such as the time of the day or the place of the hearing for instance), ``Level noise" (different decision-makers can render different decisions on the exact same case), ``Pattern noise" (different decision-makers are diversely influenced in their decision by different factors).

Interestingly, asylum decisions have been studied as a paradigmatic example of noisy decisions by \cite{noisekahneman}. Variability in asylum decisions has been previously highlighted by researchers \cite{chen_asylum_2017, dunn_early_2017, schoenholzroulette2007refugee, schoenholzroulette2007refugee, rehaag2012judicial} and in mainstream press \cite{NYTroulette}. Evidence of noise has been found in asylum decisions and experimental results show that prediction can be derived from very few features with satisfactory accuracy \cite{chen_asylum_2017, dunn_early_2017, rehaag2012judicial}. Most features used for prediction are non-substantive, non-legal, i.e. external features. There are several hypotheses to explain the apparent randomness of asylum decisions that cannot be reduced to the presence of noise or biases. While biases have been widely researched both in legal decisions and in AI, it is important that it is distinguished from noise. Biases are usually defined as systematic errors for which it is possible to identify a pattern, whether the bias is algorithmic or due to human cognition. Decision-making is often both noisy and biased at the same time. 

\subsection{Our approach and preliminary experiment}
At the intersection of Legal Judgment Prediction and Fairness studies, we aim to predict the outcome of refugee cases and identify the features that have the most impact on the decision using state-of-the-art NLP approaches (transformers) and token weight analysis. To achieve this, we use the \texttt{Ferret} library, a newly created Python library that unifies posthoc explainability methods for Transformers \cite{attanasio2023ferret}. This powerful library enables us to measure the weights of each token used as input to the prediction model. Our input data for this binary classification task consists of cases divided by paragraph, and the output is the weight assigned to each case, either 1 (granted) or 0 (denied). For a reliable evaluation, the test set is exclusively made up of cases for which we collected gold-standard annotation, while the training set contains silver-standard annotated cases. We evaluate our classifier on accuracy and macro-F1 because of the mild imbalance in classes. 

This work has 2 main objectives. The first objective is to improve transparency and fairness in decision-making and highlight potential inequalities in treatments. By analyzing the extracted features and their correlation with the decision outcome, we aim to shed light on the factors that have the most impact on the outcome of a case. This information can be used to identify any biases or disparities in the decision-making process and address them in order to improve the fairness and transparency of the system. The second objective of this work is to provide guidance for claimants and their counsels on the most important factors to consider when drafting a new application. By identifying the key features that influence the decision outcome, we can help claimants and their counsels to focus their efforts on the areas that are most likely to result in a positive outcome. This can potentially save time and resources for all parties involved, while also improving the chances of a successful outcome for the claimant.

In this section, we report the results of a preliminary experiment that we conducted on a small sample of paragraphs containing the determination of the case. We specifically focused on sentences that were flagged as belonging to the "Determination" category by our RoBERTa-based NER model. We provide an example of the output produced by our explainability method in table \ref{tab:xai table}. As expected, we found that the word "reject" had the highest weight in determining the final outcome of a case.

\begin{table*}[h!]
\centering
\begin{tabular}{|l|ccccc|}

Token	&\textbf{the} & \textbf{panel	}&\textbf{rejects	}&\textbf{the	}&\textbf{claim}\\
\hline
Partition SHAP	&-0.002938	&0.286217	&\textcolor{red}{0.437328}	&0.028208	&0.245309\\
LIME	&-0.043329	&0.169853	&0.419401	&0.13413	&0.233287\\
Gradient	&0.092172	&0.104921	&0.10768	&0.092172	&0.105733\\
Gradient x Input	&-0.016273	&0.054378	&-0.002549	&-0.016273	&0.019188\\
Integrated gradient 	&-0.120398	&-0.115244	&-0.116164	&-0.120398	&-0.114402\\
\end{tabular}
\caption{Post-hoc explainability measures per input token}
\label{tab:xai table}
\end{table*}

\subsection{Related work}
Previous  results  show  that  different  machine-learning  methods  have  been  successfully used  for  legal  prediction with  satisfactory  accuracy  levels. Previous work in legal judgment prediction mostly used linear models, with neural models being utilized more recently \cite{chalkidis-etal-2019-neural}.
 Experiments have been made comparing machine learning models (support vector machines, logR, convolutional neural networks, recurrent neural network) on legal data sets gathering decisions from the Supreme Court of the United States \cite{katz_general_2017,martin_competingsupreme_2004,ruger_supremeforecast_2004,undavia_comparative_2018} or the European Court of human rights \cite{medvedeva2020using, kaur2019convolutional}, results of which can be easily extended to asylum decisions. Similar studies have been conducted on asylum decision data sets \cite{chen_asylum_2017, dunn_early_2017, rehaag2012judicial}.
However, our work is significantly different because it relies on the raw text of the decision instead of tabular data, it is state-of-the-art in terms of machine learning and NLP as we use transformers for text classification and state-of-the-art explainability methods. Contrary to existing work, we don't measure the weights of features globally for the classifier, but we measure the weight of each token per case prediction, achieving a much high level of granularity in the analysis and therefore eliminating errors. 

\section{Limitations and ethical impact}
The ethical considerations and risks involved in this project relate to the potential negative impact on stakeholders and specifically claimants as well as judgments and legal procedures. Risks include the possibility of refusing refugee status to someone who qualifies for it, as well as granting refugee status to someone who may not fully qualify and who could be a threat to a country's security. The project must balance the need for accuracy with the risk of unintended algorithmic bias and unfairness, including procedural unfairness, that conflict with ethical AI and potentially with human rights in the context of international law.

In order to mitigate the risks of negative impact on stakeholders, judgments, and legal procedures, we plan to integrate feedback from legal professionals at every step of the process. This includes numerous important reviews of legal compliance and relevance for annotation of data and extraction of features, but also for evaluating the outcome. Our peer-review paper on a proposed methodology for human-centered legal NLP provides more details on this topic \cite{barale-2022-human}.  

\section{Conclusion and future work}
Our research concentrates on legal NLP and commences with raw data, aiming to produce explanations at the token level that exhibit greater granularity, precision, and soundness than those relying on statistical data. By utilizing state-of-the-art NLP techniques, our approach has the potential to enhance explainability and foster trust in NLP and AI-driven legal applications. As an immediate next step, we plan to continue our efforts to improve the interpretability of our transformer-based classifier. Overall, our project aims to significantly improve access to legal information for refugees and asylum seekers, while also contributing to the development of more advanced NLP applications in the legal domain.

%
%
\bibliographystyle{splncs04}
\bibliography{references}

\section{Annex}

\begin{figure*}
   \includegraphics[width=\textwidth]{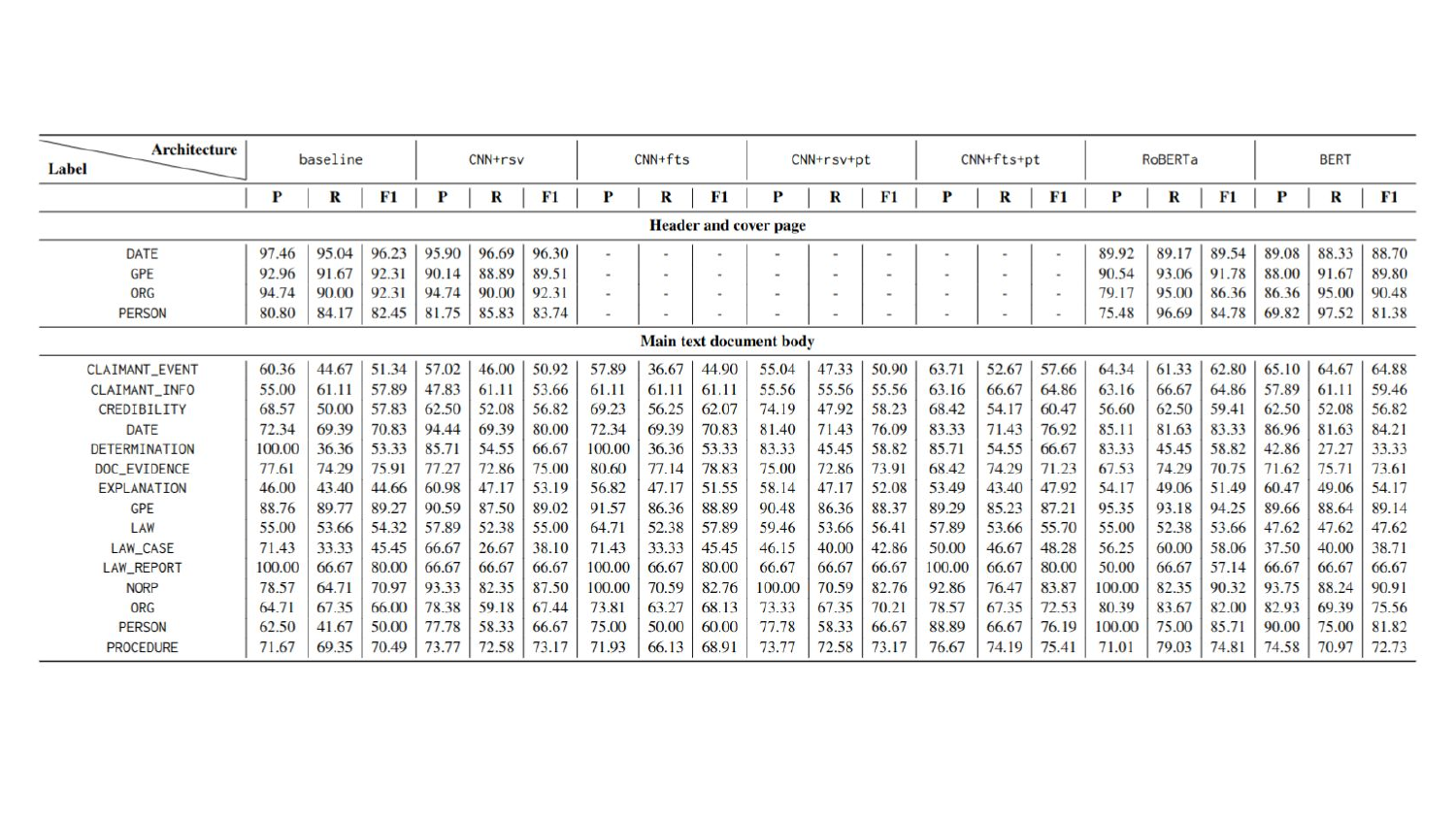} 
   \centering
   \caption{NER results -- Precision (P), Recall (R) and F1-score (in \%) on the \textit{cover page} and the \textit{main text} for seven network architectures: baseline CNN model (\texttt{baseline}), CNN model with random static vectors on \texttt{en\_core\_web\_lg} (\texttt{CNN+rsv}), CNN with fine-tuned static vectors (\texttt{CNN+fts}), CNN with random static vectors and pretraining (\texttt{CNN+rsv+pt}), CNN with fine-tuned static vectors and pretraining (\texttt{CNN+fts+pt}), \texttt{RoBERTa}-based transformer, \texttt{LegalBERT}-based transformer}
   \label{figure: table results}
\end{figure*}

\end{document}